\documentclass[sigconf]{acmart}

\AtBeginDocument{%
  \providecommand\BibTeX{{%
    \normalfont B\kern-0.5em{\scshape i\kern-0.25em b}\kern-0.8em\TeX}}}

\copyrightyear{2022}
\acmYear{2022}
\setcopyright{acmcopyright}\acmConference[WWW '22 Companion]{Companion Proceedings of the Web Conference 2022}{April 25--29, 2022}{Virtual Event, Lyon, France}
\acmBooktitle{Companion Proceedings of the Web Conference 2022 (WWW '22 Companion), April 25--29, 2022, Virtual Event, Lyon, France}
\acmPrice{15.00}
\acmDOI{10.1145/3487553.3524633}
\acmISBN{978-1-4503-9130-6/22/04}

\begin{document}

\title{A Generative Approach for Financial Causality Extraction}

\author{Tapas Nayak}
\email{tnk02.05@gmail.com}
\affiliation{%
  \institution{IIT Kharagpur}
  \country{India}
}

\author{Soumya Sharma}
\email{soumyasharma20@gmail.com}
\affiliation{%
  \institution{IIT Kharagpur}
  \country{India}
}

\author{Yash Butala}
\email{butalayash99@gmail.com}
\affiliation{%
  \institution{IIT Kharagpur}
  \country{India}
}

\author{Koustuv Dasgupta}
\email{Koustuv.x.Dasgupta@gs.com}
\affiliation{%
  \institution{Goldman Sachs}
  \country{India}
}

\author{Pawan Goyal}
\email{pawang@cse.iitkgp.ac.in}
\affiliation{%
  \institution{IIT Kharagpur}
  \country{India}
}

\author{Niloy Ganguly}
\email{niloy@cse.iitkgp.ac.in}
\affiliation{%
  \institution{IIT Kharagpur, India}
  \country{Leibniz University, Germany}
}

\renewcommand{\shortauthors}{Nayak et al.}

\begin{abstract}
  Causality represents the foremost relation between events in financial documents such as financial news articles, financial reports. Each financial causality contains a cause span and an effect span. Previous works proposed sequence labeling approaches to solve this task. But sequence labeling models find it difficult to extract multiple causalities and overlapping causalities from the text segments. In this paper, we explore a generative approach for causality extraction using the encoder-decoder framework and pointer networks. We use a causality dataset from the financial domain, \textit{FinCausal}, for our experiments and our proposed framework achieves very competitive performance on this dataset.
\end{abstract}

\ccsdesc[500]{Deep learning~Generative models}
\ccsdesc[300]{Natural language processing~Information extraction}
\ccsdesc{Event extraction~causality extraction}

\keywords{financial information extraction, financial causality extraction, generative models, pointer networks}


\maketitle

\section{Introduction}

Causality extraction from financial text is an important task for the analysis of financial documents such as financial news articles, financial reports. Previously, sequence labeling models \cite{Li2021CausalityEB,kao2020ntunlpl,becquin2020gbe} were proposed to solve this task and they assign `BIO' tags for cause and effect span to each token in the text. Although these models perform quite well in this task, they are not modeled to handle the challenges of multiple  and overlapping causalities present in the financial text. In this paper, we explore a generative approach using an encoder-decoder framework for causality extraction. We incorporate pointer networks into our decoding framework for structured prediction of the cause and effects spans in the text. The encoder-decoder approach extracts the causalities in a sequence thus the challenges of variable-length causality extraction and overlapping causalities extraction are solved. The pointer network-based decoding identifies the cause and effect spans using the start and end positional index in the text. So the cause and effect spans of different lengths are modeled uniformly in this approach. We use \textit{FinCausal} \cite{mariko2020financial}, a dataset containing text segments from financial news articles, for our experiments. Our proposed model achieves very competitive performance in this dataset. We release our code and data for future research at \url{https://github.com/nayakt/CEPN}.

\section{The \textbf{CEPN} Framework}
\begin{figure}[t]
\centering
\includegraphics[scale=0.3]{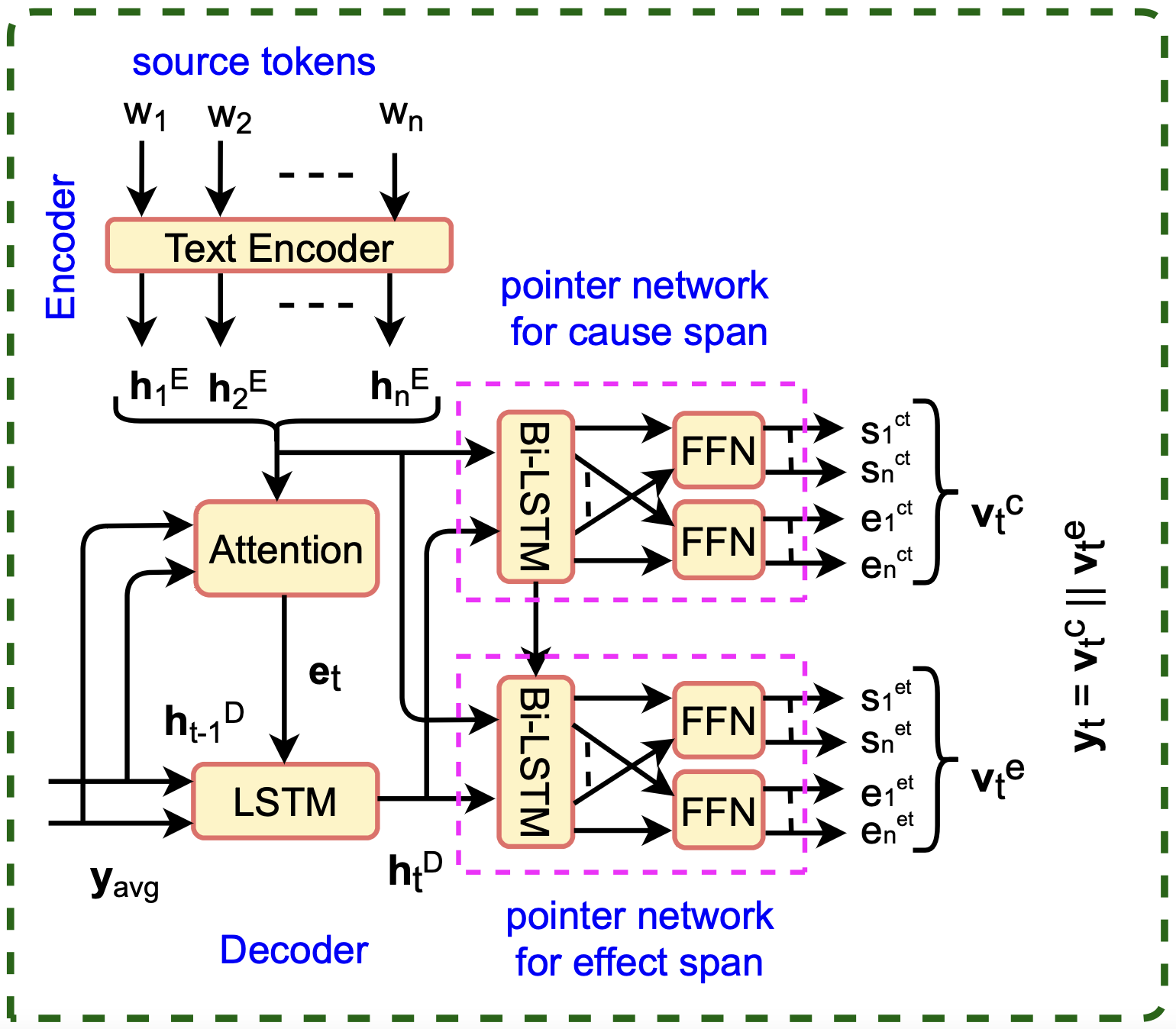}
\caption{The architecture of CEPN framework.}
\label{fig:ptrnet}
\end{figure}

We present \textbf{CEPN}, a \textbf{C}ausality \textbf{E}xtraction framework using \textbf{P}ointer \textbf{N}etwork-based encoder-decoder model. To formally define this task, given a text segment $S=\{w_1, w_2, ..., w_n\}$ with $n$ tokens, the goal is to extract a set of causalities $T=\{y_1, y_2, ...., y_{m-1}\}$ where $y_t=(c_s^t, c_e^t, e_s^t, e_e^t)$ is the $t^{th}$ causality, $m-1$ is the number of causalities. $c_s^t$, $c_e^t$, $e_s^t$, and $e_e^t$ represent the positions of the start and end tokens of the cause and effect span of the $t^{th}$ causality in the text segment $S$, respectively. In Fig \ref{fig:ptrnet}, we give an overview of our proposed model which is inspired from the models proposed in similar structure prediction tasks such as joint entity-relation extraction \citep{nayak2020ptrnet} and aspect-sentiment triplet extraction \citep{Mukherjee2021PASTE}.

We use a pre-trained BERT model \cite{Devlin2019BERTPO} to encode the source text. We concatenate the part-of-speech tag embeddings of the tokens with the BERT vectors to obtain the encoder hidden states $\mathbf{h}_i^E \in \mathbb{R}^{d_h}$. We add a special token `[unused0]' at the front of the text and use its positional index `$0$' to stop the decoding process. We use BERT\_Base\_Cased of dimension $768$ and BERT\_Large\_Cased of dimension $1,024$ model for encoding, and we refer  them as CEPN\_Base and CEPN\_Large. We set the dimension of the part-of-speech tag embeddings at $32$, and initialize them randomly. The hidden dimension $d_h$ of encoder for CEPN\_Base and CEPN\_Large model is $768+32=800$ and $1,024+32=1,056$, respectively.

We consider the causalities as a sequence $T=\{y_1, y_2,....,y_{m-1}\}$.  We use two special tuples to model this task as a sequence generation problem. We start the sequence generation process with a special tuple $y_0$, and we mark the end of this generation process with another special tuple, $y_m=(0, -1, -1, -1)$. Here, only $c_s^t=0$ (corresponds to [unused0] token) is enough to mark the end of the sequence generation process, and we ignore the other indexes of value $-1$. In the decoder, we use an LSTM cell for generating the causality sequence. We set the hidden dimension of this LSTM cell at $d_h$ that is the same as the hidden dimension of the encoder vectors. We pass the encoder context representation, $\mathbf{e}_t \in \mathbb{R}^{d_h}$, and the average of the vectors of already generated causalities, $\mathbf{y}_{avg}$ to this LSTM cell to generate the next hidden state, $\mathbf{h}_t^D \in \mathbb{R}^{d_h}$. We use two pointer networks for identifying the cause and effect span of the causalities. Each pointer network consists of a BiLSTM layer with hidden dimension $d_p$ and two feed-forward layers (FFN) of dimension $d_p \times 1$ with Softmax activation. We set $d_p=d_h$ as the hidden dimension of the BiLSTM layer in the pointer networks. The output of this FFN/Softmax layer is a scalar score corresponding to each token in the source text. These scalar scores represent the probability distribution of the source tokens being the start or end token of the cause/effect span. We can extract either of the cause or effect spans first, then use the vector representation of that span to extract the other span of a causality. In one version, we extract cause span first, we refer to them using `CF' such as CEPN\_Base\_CF. We experiment with another version where we extract the effect span first, and we refer to them using `EF' such as CEPN\_Base\_EF.

\section{Experiments}

We use the \textit{FinCausal} 2020\footnote{http://wp.lancs.ac.uk/cfie/fincausal2020/} and \textit{FinCausal} 2021\footnote{http://wp.lancs.ac.uk/cfie/fincausal2021/} datasets for our experiments. The text segments in these two datasets contain a maximum of three sentences and the cause and effect span of causality can appear in different sentences in the segment. The number of average tokens in the text segments is $\sim$ 40 and in the cause and effect span is $\sim$ 17-18. As the gold-label causalities of the blind test split are not publicly available, we do a 5-fold cross-validation on the training datasets for experiments. We compare our model against the previous SOTA models such as UPB \cite{ionescu2020upb}, DOMINO \cite{chakravarthy2020domino}, PAMNet \cite{szanto2020prosperamnet}, GBe \cite{becquin2020gbe}, NTUNLPL \cite{kao2020ntunlpl}. We report token-level F1 score and exact-match F1 score for comparison.



We include the results on \textit{FinCausal} 2020 and \textit{FinCausal} 2021 datasets in Table \ref{tab:results}. We see that our CF and EF versions of the model perform almost equally on both of these datasets. Compare to GBe\_Large model, our CEPN\_Large\_EF model achieves 1.4\% higher token-level F1 score and 3.3\% higher exact match-based F1 score on the \textit{FinCausal} 2020 dataset. On the \textit{FinCausal} 2021 dataset, compare to GBe\_Large model, our CEPN\_Large\_CF model achieves 0.9\% higher token-level F1 score and 1.4\% higher exact match-based F1 score in this dataset. We also perform a statistical significance test (two-tailed and paired) between CEPN\_Large\_CF model and GBe\_Large model on the \textit{FinCausal} 2020 and 2021 datasets and find that our model is statistically significant with $p < 0.001$ (achieves a 1.1\% higher mean token-level F1 score on both datasets).

\begin{table}[ht]
\small
\centering
\begin{tabular}{l|cc|cc}
\hline
                    & \multicolumn{2}{c|}{FinCausal 2020}                        & \multicolumn{2}{c}{FinCausal 2021}     \\ \hline
Model               & Token\_F1     & EM\_F1    & Token\_F1 & EM\_F1 \\ \hline
UPB\_Base \cite{ionescu2020upb}       & 0.689         & -         &   -        &    -    \\ 
DOMINO\_Base \cite{chakravarthy2020domino}    & 0.796         & -         &   -        &   -     \\ 
PAMNet\_Large* \cite{szanto2020prosperamnet} & 0.767   & 0.485  & 0.776   & 0.533    \\ 
GBe\_Large* \cite{becquin2020gbe}    & 0.860   & 0.710  & 0.883  & 0.761    \\ 
NTUNLPL\_Base \cite{kao2020ntunlpl} & 0.869  & -                 &    -       &    -    \\ \hline
CEPN\_Base\_CF      & 0.866  & 0.733  & 0.885  & 0.769          \\ 
CEPN\_Base\_EF     & 0.857  & 0.722  & 0.883  & 0.761         \\ 
CEPN\_Large\_CF     & 0.870  & 0.739  & \textbf{0.892}  & \textbf{0.775}          \\ 
CEPN\_Large\_EF    & \textbf{0.874}  & \textbf{0.743}  & 0.891  & 0.767       \\ \hline
\end{tabular}
\vspace{2mm}
\caption{Performance comparison of CEPN model on the \textit{FinCausal} datasets against the previous SOTA models. * marked baseline scores are reproduced by us. The remaining baseline scores are taken from their papers. `Base' and `Large' refer to the use of BERT\_Base and BERT\_Large models for source encoding. We report the average of the 5-folds in a cross-validation setting.}
\label{tab:results}
\end{table}

\section{Conclusion}

In this paper, we explore a generative approach for causality extraction in financial text using the encoder-decoder framework and pointer networks. Our model is designed to handle different challenges in this task such as extracting unknown variable-length causalities, identifying text with no causality, and extracting overlapping causalities from the text in an end-to-end architecture. Experimental results on the \textit{FinCausal} datasets show the effectiveness of our proposed framework on financial causality extraction.

\begin{acks}
This research was partially supported by Goldman Sachs under the research grant FTHS (FinTalk: Research towards creating a platform for highlight generation and summarization of financial documents while taking into account user feedback).
\end{acks}

\bibliographystyle{ACM-Reference-Format}
\bibliography{main}










\end{document}